\def\year{2021}\relax
\documentclass[letterpaper]{article}
\usepackage{aaai21} 
\usepackage{times} 
\usepackage{helvet} 
\usepackage{courier} 
\usepackage[hyphens]{url} 
\usepackage{graphicx} 
\usepackage{graphicx}  
\usepackage{natbib}
\usepackage{caption}
\usepackage{verbatim}
\usepackage{amsmath}
\usepackage{amssymb}
\usepackage{algorithm}
\usepackage{algpseudocode}
\usepackage{subfig}

\begin{document}

%
\title{Comparison Lift: Bandit-based Experimentation System for Online Advertising}
\author{Tong Geng$^1$, Xiliang Lin$^1$, Harikesh S. Nair$^{1,2}$, Jun Hao$^1$, Bin Xiang$^1$, Shurui Fan$^1$ \\
{{tong.geng,xiliang.lin,haojun,xiangbin1,fanshurui}@jd.com, harikesh.nair@stanford.edu}\\
{$^1$JD.com   }
{    $^2$Stanford University}
}
\maketitle
\begin{abstract}
\begin{footnotesize}\texttt{Comparison Lift}\end{footnotesize} is an experimentation-as-a-service (EaaS) application for testing online advertising audiences and creatives at JD.com. Unlike many other EaaS tools that focus primarily on fixed sample A/B testing, \begin{footnotesize}\texttt{Comparison Lift}\end{footnotesize} deploys a custom bandit-based experimentation algorithm. The advantages of the bandit-based approach are two-fold. First, it aligns the randomization induced in the test with the advertiser's goals from testing. Second, by adapting experimental design to information acquired during the test, it reduces substantially the cost of experimentation to the advertiser. Since launch in May 2019,  \begin{footnotesize}\texttt{Comparison Lift}\end{footnotesize} has been utilized in over 1,500 experiments. We estimate that utilization of the product has helped increase click-through rates of participating advertising campaigns by 46\% on average. We estimate that the adaptive design in the product has generated 27\% more clicks on average during testing compared to a fixed sample A/B design. Both suggest significant value generation and cost savings to advertisers from the product.
\end{abstract}

\section{Introduction}

Compared to their offline, pre-internet age brethren, online, internet-enabled digital marketing campaigns are extremely complex. One source of complexity is targeting. Compared to age and gender-based targeting that are common in say television ad-markets, digital marketing campaigns offer targetability of users on the basis of a large variety of demographic, contextual and behavioral features. For instance, on e-commerce platforms $-$ increasingly large-scale facilitators of online advertising $-$ target audiences can be defined on the basis of very flexible criteria such as ``male iPhone users who bought Nike shoes in the last 6 months and browsed the shoes category page in the last 3 days.'' Any of these parameters could be changed to obtain a different target audience, producing a large number of potential audiences that could be addressed by a campaign. Another source of complexity is the type of media that can be used in digital advertising campaigns. A wide variety of text, images, videos, both short and long form, as well as landing pages are possible, with different types of variants or combinations targeted to different sub-populations. Finally several campaigns tend to be omnichannel, involving search and display advertising and spanning ad inventory across app, browser, PC and mobile modalities. While the array of options has opened up a golden age of possibilities for digital marketing, it has also produced challenges for advertisers and marketers in designing effective campaigns by presenting them with a bewildering number of feasible design combinations. A principled approach to campaign design that is data-driven, automated, and based on proper foundations of causality and incremental response has therefore become key to effective marketing strategy. The product described in this paper, \begin{footnotesize}\texttt{Comparison Lift}\end{footnotesize}, is a self-serve, advertiser-facing product we designed and deployed to deliver on this need for the advertising business of JD.com, a large e-commerce company in China.

\begin{footnotesize}\texttt{Comparison Lift}\end{footnotesize} works by leveraging randomized controlled trials to allow advertisers to directly field-test various target audiences, creatives, and their combinations against one another. The target audiences can be specified flexibly, and the creatives can comprise most type of media (images, text, videos), so the scope of tests facilitated by the product is broad.

The product has three distinguishing features. First, it implements randomized controlled trials using adaptive, bandit-based designs. To understand the usefulness of this, note that a typical solution to the problem of experimentally finding the best creative amongst a set of possible variants is to use an ``A/B/n'' design, sometimes referred to as a ``split-test''. This design keeps the sample traffic splits across creatives constant as the test progresses. Therefore, both good and bad creatives will be allocated the same amount of traffic. In contrast, the bandit-based system implemented here adapts the traffic allocation dynamically, reducing the allocation of experimental traffic to creatives that are learned to perform poorly, thus reducing the cost of experimentation significantly relative to non-adaptive designs. In addition, by basing the allocation of experimental traffic on maximizing the payoff to the advertiser from the best-discovered creatives, the algorithm aligns the basis for test randomization directly with the goals of the advertiser from testing.

A second feature is it leverages new algorithms that handle complex target audiences so as to test both target audiences and creatives simultaneously and to identify best matches between creatives and audiences. To understand this point, note that when target audiences have to be evaluated in addition to creatives, simple extensions of ``A/B/n'' designs face difficulties in evaluating audiences that overlap with each other. Without adjustments, data is under-utilized or biases are induced because the samples of users from various target audiences collected in the test become non-representative of the platform population. See \cite{Geng2020}. The algorithm utilized in \begin{footnotesize}\texttt{Comparison Lift}\end{footnotesize} handles these issues in a seamless way.

A third feature is that the Bayesian inference implied by the algorithm is more resilient than comparable fixed sample frequentist approaches, in flexibly allowing advertisers to monitor the progress of experiments and to stop and re-start them. This increases the flexibility of the product and allows advertisers more control in their testing plans, which is desirable.

These aspects make \begin{footnotesize}\texttt{Comparison Lift}\end{footnotesize} a novel product. In contrast, to our knowledge, other ad-experimentation products currently available in industry use either non-adaptive ``split-test'' schemes, or when adaptive, are designed to test different creatives on a single targeting audience, or are used for internal experiments rather than as an external-facing product for advertisers.\footnote{\begin{footnotesize}\texttt{Optimize}\end{footnotesize} at Google, \begin{footnotesize}\texttt{A/B tests}\end{footnotesize} at Facebook, \begin{footnotesize}\texttt{Experiments Learning Center}\end{footnotesize} at Amazon, and \begin{footnotesize}\texttt{Split-Sample Test}\end{footnotesize} at Tencent are industry tools to help advertisers run experiments.}

The product was deployed in May 2019.  By the end of June, 2020, 1,547 experiments had run on the product. A typical experiment runs for 5 days, tests 2.4 creatives and 3.8 target audiences. To measure the business impact and value generated by the product, we construct two metrics. The first measures the value of experimentation to the advertiser. To do this, we compare click-through rates (henceforth ``CTR''s) for the best option discovered by the advertiser via the experiment, with what she would have obtained counterfactually without the experiment. The focus on CTR is particularly relevant on an e-commerce platform such as JD.com, as clicks on most ads drive the user directly to the SKU-detail page on the platform, and thereby generate awareness and visitation of the product, which are key goals of the advertiser. We estimate the best discovered options yields 46\% more CTRs on average to advertisers. The second metric measures the value of the adaptive design by comparing the number of clicks generated from the adaptive design with the counterfactual number of clicks generated by a non-adaptive experimental design. We estimate the utilized design generates 27\% more clicks on average than a non-adaptive design, representing a net lowering of advertisers' experimental costs. Both suggest the product has generated a significant reduction in uncertainty for advertisers and helped improve campaign design at much lower costs than typical approaches.

\section{Algorithm}
The problem addressed is as follows. An advertiser designing a campaign wants to pick, from a set of possible target audiences and creatives, a \emph{creative-target audience combination} that provides her the highest expected payoff in a campaign. We would like to design an experiment, embedded within a product, to find the best creative-target audience combination for the advertiser while minimizing her costs of experimentation.

\begin{footnotesize}\texttt{Comparison Lift}\end{footnotesize} incorporates a new algorithm described in \cite{Geng2020} that addresses these issues. It has two broad steps. In step one, it splits the compared target audiences (henceforth ``\textit{TA}''s) into disjoint audience sub-populations (henceforth ``\textit{DA}"s), so the set of \emph{DA}s fully span the set of \emph{TA}s. In step two, we train a bandit with the creatives as arms, the payoffs to the advertiser as rewards, and the \emph{DA}s, rather than the \emph{TA}s as the contexts. As the test progresses, we aggregate over all \emph{DA}s that correspond to each \emph{TA} to adaptively learn the best creative-\emph{TA} match (henceforth ``\emph{C-TA}''). In essence, we learn an optimal creative allocation policy at the disjoint sub-population level, while making progress towards the test goal at the \emph{TA} level. Because the \emph{DA}s have no overlap, each user can be mapped to a distinct \emph{DA}, resolving an assignment problem that arises in alternative designs that directly compare overlapping \emph{TA}s. Because all \emph{DA}s that map to a \emph{TA} help inform the value of that \emph{TA}, learning is accelerated. Tailoring the bandit's policy to a more finely specified context $-$ i.e., the \emph{DA} $-$ allows it to match the creative to the user's tastes more finely, thereby improving payoffs and reducing expected regret, while delivering on the goal of assessing the best combination at the level of a more aggregated audience. The adaptive nature of the test ensures the traffic is allocated in a way that reduces the cost to the advertiser from running the test, because creatives that are learned to have low value early are allocated lesser traffic within each \emph{DA} as the test progresses. The overall algorithm is implemented as a contextual Thompson Sampler (henceforth ``TS''; see \cite{russo2018} for an overview).

\subsubsection{Details}
The advertiser provides as input $\mathbb{K} = \{1,..,K\}$ possible \emph{TA}s and $\mathbb{R} = \{1,..,R\}$ creatives she wants to evaluate for her campaign. The $K$ \emph{TA}s are partitioned into a set $\mathbb{J} = \{1,..,J\}$ of $J$ \emph{DA}s. When a user $i$ arrives at the platform, we categorize the user to a context based on his features, i.e., $i\in DA(j) \text{ if } i\text{'s features match the definition of } j$, 
where $DA(j)$ denotes the set of users in DA $j$. The context determines which creative $r\in\mathbb{R}$ is displayed to the user.

To set up the model, let $y_{irj}$ be an indicator for whether $i$ clicks on creative $r$ when its displayed. We model $y_{irj}$ in a Bayesian framework, and let,
\begin{equation}
y_{irj}\sim \texttt{Ber}(\theta_{rj}); \text{and,  } \theta_{rj}\sim\texttt{Beta}(\alpha_{rj},\beta_{rj}).
\label{eq:outcome-and-hyper}
\end{equation}

\noindent where $\theta_{rj}$ is the CTR, and $\Omega_{rj}\equiv(\alpha_{rj},\beta_{rj})$ are the hyper-parameters governing the distribution of $\theta_{rj}$. The use of a Beta prior for a Bernoulli distributed outcome ($y$) is helpful as it is a conjugate to the likelihood, enabling fast updating of the posterior as the experiment progresses. 

The payoff to the advertiser from the ad-impression is defined as $\pi_{irj}$ = $\gamma\cdot y_{irj}-b_{irj}$, where $\gamma$ is a factor that converts clicks to monetary units, and $b_{irj}$ denotes the cost of displaying creative $r$ to a user $i$ of context $j$.\footnote{$\gamma$ may be determined from prior estimation or advertisers' judgment of the value attached to users' actions. $\gamma$ is pre-computed and held fixed during the test. $\bar{b}_{rj}$ and $\hat{p}(j|k)$ (defined later) can be pre-computed outside of the test from historical data and held fixed during the test, or inferred during the test using a simple bin estimator that computes these as averages over the observed cost and user contexts data.}

Combining this payoff with equation (\ref{eq:outcome-and-hyper}) implies the expected payoff of each creative-disjoint sub-population combination (henceforth ``\emph{C-DA}'') is $\mu_{rj}^{\pi}(\theta_{rj})$ = $\mathbb{E}[\pi_{irj}]$ = $\gamma\mathbb{E}[y_{irj}]-\mathbb{E}[b_{irj}]$ = $\gamma\theta_{rj}-\bar{b}_{rj}$, $\forall r\in\mathbb{R},j\in\mathbb{J}$, 
where $\bar{b}_{rj}$ is the average cost of showing  $r$ to users in $DA(j)$.

The TS aims to find an optimal policy $g(j):\mathbb{J}\rightarrow\mathbb{R}$ that allocates the creative with the maximum expected payoff to a user with context $j$. To make clear how the TS updates parameters, we add the index $t$ for batch. Before the test starts, $t=1$, we set diffuse priors and let $\alpha_{rj,t=1}=1,\beta_{rj,t=1}=1,\forall r \in \mathbb{R}, j \in \mathbb{J}$. This implies the prior probability of clicking, $\theta_{rj,t=1},\forall r \in \mathbb{R}, j \in \mathbb{J}$ is uniformly distributed between 0\% and 100\%.

In batch $t$, $N_{t}$ users arrive. The TS displays creatives to these users dynamically, by randomly allocating each creative according to the posterior probability each creative offers the highest expected payoffs given a user's context. Let $r_{jt}^{*}$ denote the creative with highest expected payoff within context $j$ given the posterior at the beginning of batch $t$. The probability $r_{jt}^{*}$ is $r$ is $ w_{rjt}=
    Pr[\mu_{rj}^{\pi}(\theta_{rjt})=\max\limits_{r \in \mathbb{R}} (\mu_{rj}^{\pi}(\theta_{rjt}))|\vec{\alpha}_{jt},\vec{\beta}_{jt}]$. Here, $\vec{\alpha}_{jt}$ = $[\alpha_{1jt}, \dots, \alpha_{Rjt}]'$ and $\vec{\beta}_{jt}$ = $[\beta_{1jt}, \dots, \beta_{Rjt}]'$ are the parameters of the posterior distribution of $\vec{\theta}_{jt}$ = $[\theta_{1jt},\dots,\theta_{Rjt}]'$.

We update all parameters at the end of processing the batch, after the outcomes for all users in the batch is observed. We compute the sum of binary outcomes for each \emph{C-DA} combination as  $s_{rjt}=\sum_{i=1}^{n_{rjt}}y_{irjt}$, $\forall r \in \mathbb{R},j \in \mathbb{J}$, where $n_{rjt}$ is the number of users with context $j$ allocated to creative $r$ in batch $t$. Then, we update parameters as $\vec{\alpha}_{j(t+1)}=\vec{\alpha}_{jt}+\vec{s}_{jt}$ and $\vec{\beta}_{j(t+1)}=\vec{\beta}_{jt}+\vec{n}_{jt}-\vec{s}_{jt}$, $\forall j \in \mathbb{J}$,  
where $\vec{s}_{jt} = [s_{1jt},\dots,s_{Rjt}]'$, and $\vec{n}_{jt} = [n_{1jt},\dots,n_{Rjt}]'$. Then, we enter batch $t+1$, and use $\vec{\alpha}_{j(t+1)}$ and $\vec{\beta}_{j(t+1)}$ as the posterior parameters to allocate creatives in $t+1$. 

While the bandit contextually learns the best \emph{C-DA} combination in this manner, in parallel, we compute the expected payoff of each \emph{C-TA} combination by appropriately aggregating the payoffs of corresponding \emph{C-DA} combinations in the experiment. To do this, we aggregate across all \emph{C-DA}s associated with \emph{C-TA} combination $(r,k)$ to obtain $\lambda_{rkt}=\sum_{j\in\mathcal{O}(k)}\theta_{rjt}\cdot\hat{p}(j|k)$. Here, $\lambda_{rkt}$ is interpreted as the CTR of a typical user from TA($k$) for creative $r$; $\hat{p}(j|k)$ is the probability (in the platform population) that a user from $TA(k)$ has context $j$; and $\mathcal{O}(k)$ is the set of disjoint sub-populations ($j$s) whose associated \emph{DA($j$)}s are subsets of $TA(k)$. Each $\lambda_{rkt}$ implies a corresponding expected payoff to the advertiser from displaying creative $r$ to a user from TA($k$) obtained as, $\omega_{rkt}^{\pi}(\lambda_{rkt})=\gamma\lambda_{rkt}-\bar{b}_{rk}\ensuremath{,}\forall r\in\mathbb{R},k\in\mathbb{K}$.  Here, $\bar{b}_{rk}$ is the average cost for showing creative $r$ to target audience $k$, which can be obtained by aggregating $\bar{b}_{rj}$ over $j\in\mathcal{O}(k)$ in the same manner as $\lambda_{rkt}$s. 

\begin{figure}
\centering
\includegraphics[width=0.35\textwidth]{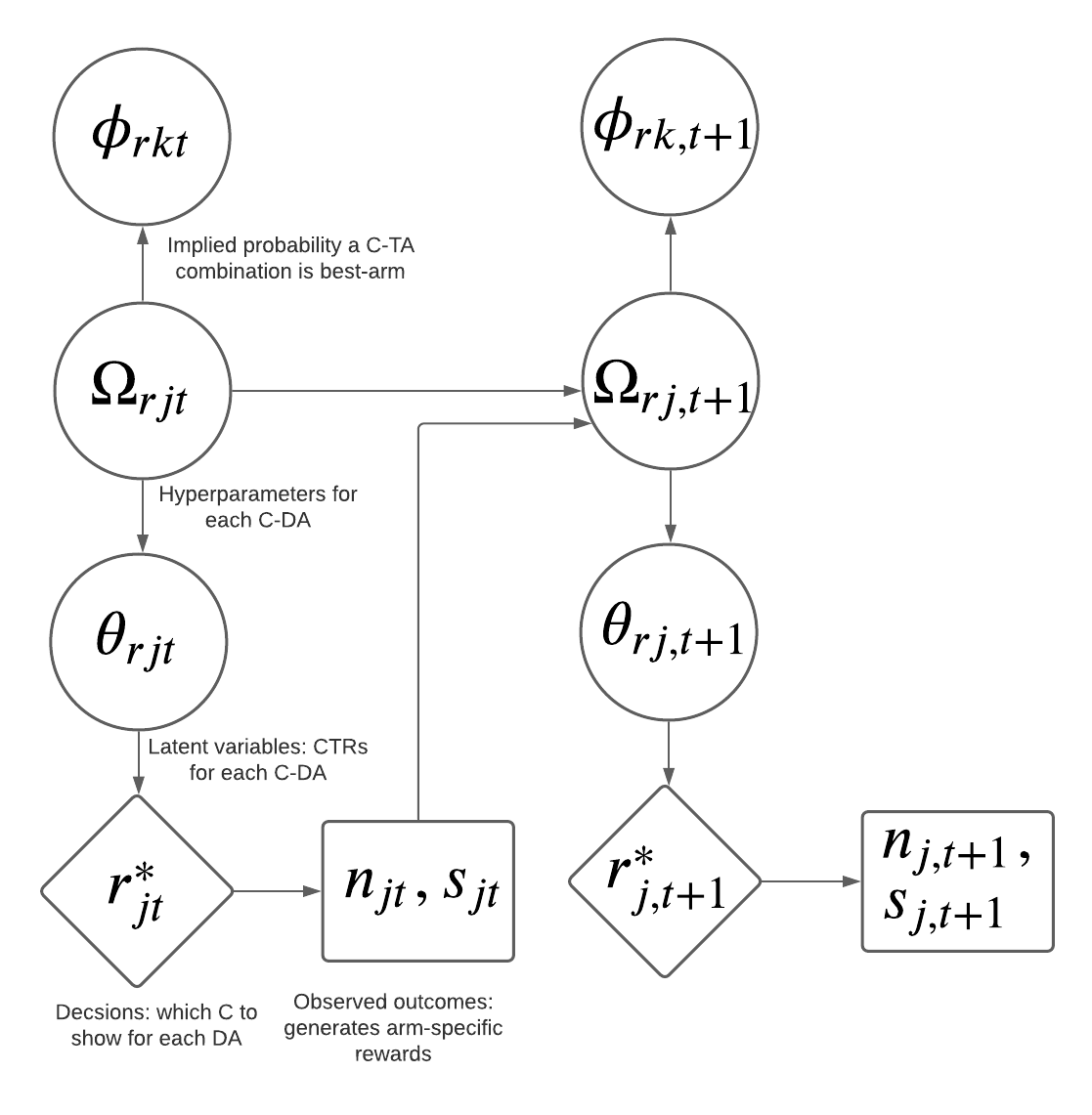}
\caption{Schematic of Algorithm}
\label{fig:DAG}
\end{figure}

The progress towards the goal is measured by computing $\phi_{rkt}$, the posterior probability in batch $t$ that a \textit{C-TA} combination $(r,k)$ is best, i.e.,  $\phi_{rkt}$ = $Pr[\omega_{rkt}^{\pi}(\lambda_{rkt})$ = $\max\limits _{r\in\mathbb{R},k\in\mathbb{K}}(\omega_{rkt}^{\pi}(\lambda_{rkt})|\vec{\alpha}_{.t},\vec{\beta}_{.t})]$. We can approximate $\phi_{rkt}$ using using Monte Carlo sampling. To do this, notice the posterior distribution of $\theta_{rjt}$s from the TS induces a distribution of $\lambda_{rkt}$s. To sample from this distribution, for each batch, we make $H$ draws $\ensuremath{\theta_{rjt}^{\left(h\right)}},h=1,..,H$ from the current posteriors $\texttt{Beta}(\alpha_{rjt}, \beta_{rjt})$, and use them to construct $H$  corresponding values of $\ensuremath{\lambda_{rkt}^{\left(h\right)}},h=1,..,H$. For each such $\lambda_{rkt}^{\left(h\right)}$,  we compute $\omega_{rkt}^{\pi}(\lambda_{rkt}^{\left(h\right)})=\gamma\lambda_{rkt}^{\left(h\right)}-\bar{b}_{rk}$, $\forall r\in\mathbb{R},k\in\mathbb{K}$. Then, we estimate $\phi_{rkt}$ as the proportion of draws in which combination $(r,k)$ is the best, i.e., $\phi_{rkt}=\frac{1}{H}\sum_{h=1}^{H}I[\omega_{rkt}^{\pi}(\lambda_{rkt}^{\left(h\right)})=\max\limits _{r\in\mathbb{R},k\in\mathbb{K}}(\omega_{rkt}^{\pi}(\lambda_{rkt}^{\left(h\right)}))|\vec{\alpha}_{.t},\vec{\beta}_{.t}]$, where $I(.)$ is the indicator function.  Figure \ref{fig:DAG} provides a schematic of this procedure.

Once $\phi_{rkt}$ for any one $(r,k)$ combination crosses a pre-specified probability threshold, the advertiser is notified that the experiment has identified a best-performing \emph{C-TA} combination. The advertiser is free to stop the experiment at this point, let it run further, or apply the best chosen configuration to a new campaign.\footnote{\cite{Geng2020} discuss how to develop a formal stopping criteria for the experiment based on regret performance, and presents more details and benchmarks.} 

\subsubsection{Case Study}
\begin{figure}
\centering
\includegraphics[width=0.35\textwidth]{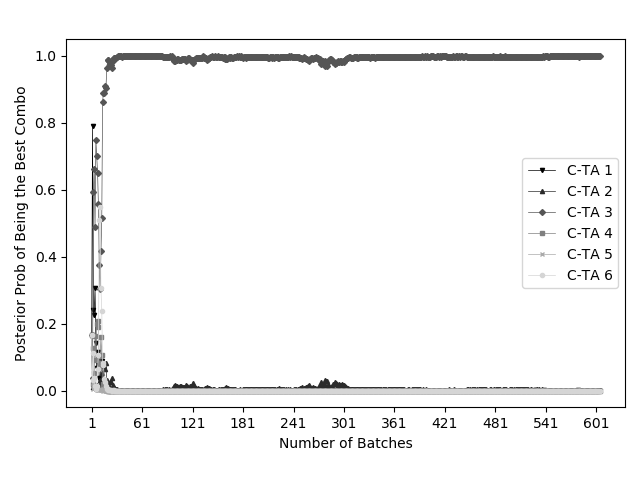}
\caption{Case-study: Prob(\emph{C-TA}) combo is best}
\label{fig:case_sim}
\end{figure}

To obtain a feel for how the algorithm works in practice, we discuss a case-study based on a test implemented by Samsung, a large cellphone manufacturer and advertiser on JD.com. Samsung seeks to find the best \emph{C-TA} combination across 2 candidate \emph{TA}s and 3 creatives for a campaign it is considering. The 2 \emph{TA}s overlap, resulting in 3 \emph{DA}s. Figure \ref{fig:case_sim} shows the probability that each \emph{C-TA} combination is estimated to be the best as the test progresses. The 6 possible combinations are shown in different shades and markers. Within the initial 50 batches, the algorithm identifies combinations 1,4,5 as inferior and focuses on exploring the other 3 combinations. Then, combination 3 starts to dominate the others and is finally identified as the best. Most of the traffic during the test is allocated to combination 3, so the advertiser does not unnecessarily waste resources on assessing those learned to be inferior early on.

\begin{figure}
\centering
\includegraphics[width=0.35\textwidth]{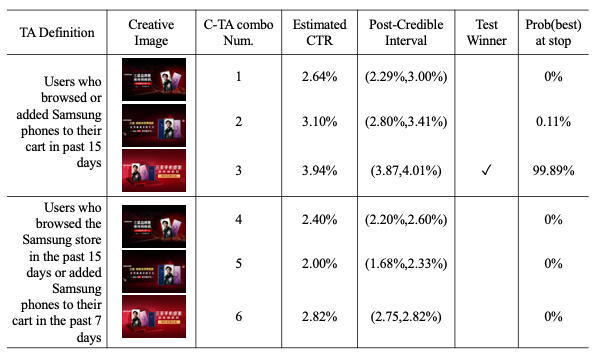}
\caption{Results from Case Study}
\label{fig:case_report}
\end{figure}

Figure \ref{fig:case_report} shows results at the end of the test. We estimate a high posterior probability on combination 3 being the best (99.91\%). The difference in CTRs between the best and worst combinations is indicative of the value of the test in resolving advertiser uncertainty. In this test, the CTR of the best combination (3.94\%) is 97\% higher than the worst (2.0\%), suggesting significant uncertainty reduction. 

To obtain a sense of value generated for the advertiser, we develop two metrics. One metric which we call \emph{value of experimentation} compares the performance of the best discovered combination to what the advertiser would have obtained counterfactually in the absence of the test. We assume that in the absence of the test, the advertiser would pick one of the 6 tested options. Assigning equal probability to each of these possibilities, we compute the metric as the CTR of the best option divided by the average of the CTRs of the tested combinations. For this test, this metric is about 1.4, suggesting the test generates about 40\% more clicks to the advertiser compared to implementing her campaign without running the test, \emph{ceteris paribus}.

A second metric, which we call \emph{value of adaptive design} helps assess the value of the experimental design. This compares the clicks factually generated in the test, with what would be generated counterfactually if the test were implemented using a non-adaptive ``A/B/n'' design. We estimate the number of clicks generated in the counterfactual regime by simulating an equal allocation scheme that allocates the total impressions obtained in the test equally to each tested creative. To do this, we take users of each \emph{DA} in the test, and divide them equally across the creatives. Then, we apply the CTRs estimated from the test for each \emph{C-DA} combination, to obtain an estimate of the total clicks that would be generated under this alternative allocation. Dividing the observed number of clicks by the simulated total clicks gives the metric. For this case-study, this metric is about 1.17, suggesting a 17\% reduction in the opportunity cost to the advertiser of serving inferior creatives in the test, \emph{ceteris paribus}. Later in the paper, we compute both metrics across a broader set of tests to assess how these results extend more generally.

\section{Deployment}
This section presents an overview of the product development and deployment effort, along with a discussion of how the product is engineered and implemented.
\subsubsection{Development and Launch}
The product was developed over a period of 5 months by a team of 3 research scientists, 5 engineers, a User Interface (``UI'') designer, product manager, project manager and salesperson. 
In the first step of development $-$ comprising roughly 2.5 months $-$ the science team developed the algorithm and created a working prototype that demonstrated its efficacy and performance. Once the algorithm was approved for production, the science team worked with the engineering team to encode the algorithm into the complex ad-serving system of JD.com. This required building an infrastructure to ensure the experiment does not interrupt normal ad-serving flow or induce unviable latency into it, while ensuring accurate real-time updating of bandit posteriors and data tracking (described in more detail below). Upon completion of this step, several pilots were run with advertisers to assess algorithm performance. Aspects that were checked included whether the collected data updated model parameters correctly, whether the bandit indeed allocated traffic reflecting updated parameters, and whether the test report produced correct results and credible intervals. After this process, the UI team designed an interface that allows advertisers to run \begin{footnotesize}\texttt{Comparison Lift}\end{footnotesize} tests, as well as a portal for them to monitor test progress and obtain reports. Subsequently, data pipelines linking the data and the UI were built, and additional testing implemented to identify potential loopholes in the full system. 

The product was launched in April 2019 and opened to a whitelisted set of advertisers. After a one-month period, during which product usage was monitored and advertiser feedback was received and incorporated, the product was opened to all advertisers in May 2019. As part of launch, several manuals and collateral were developed to help advertisers understand the algorithm and the value of the product (see \url{https://jzt.jd.com/study/tool/2110.jhtml}), as well as how to set up tests (see \url{https://jzt.jd.com/study/yhdcintroduce/1815.jhtml}). After launch, product performance was monitored and advertiser feedback incorporated continuously. Over time, the product was expanded to cover more ad-inventory on the JD platform. Throughout this time-frame, a product manager was responsible for managing the productization process, a project manager for coordinating scientists and engineers involved in the deployment, and a salesperson for communicating with and collecting feedback from advertisers. 

\subsubsection{Product: User Interface and Experiment Set-up}
\begin{figure}
	\centering
	\includegraphics[width=0.45\textwidth]{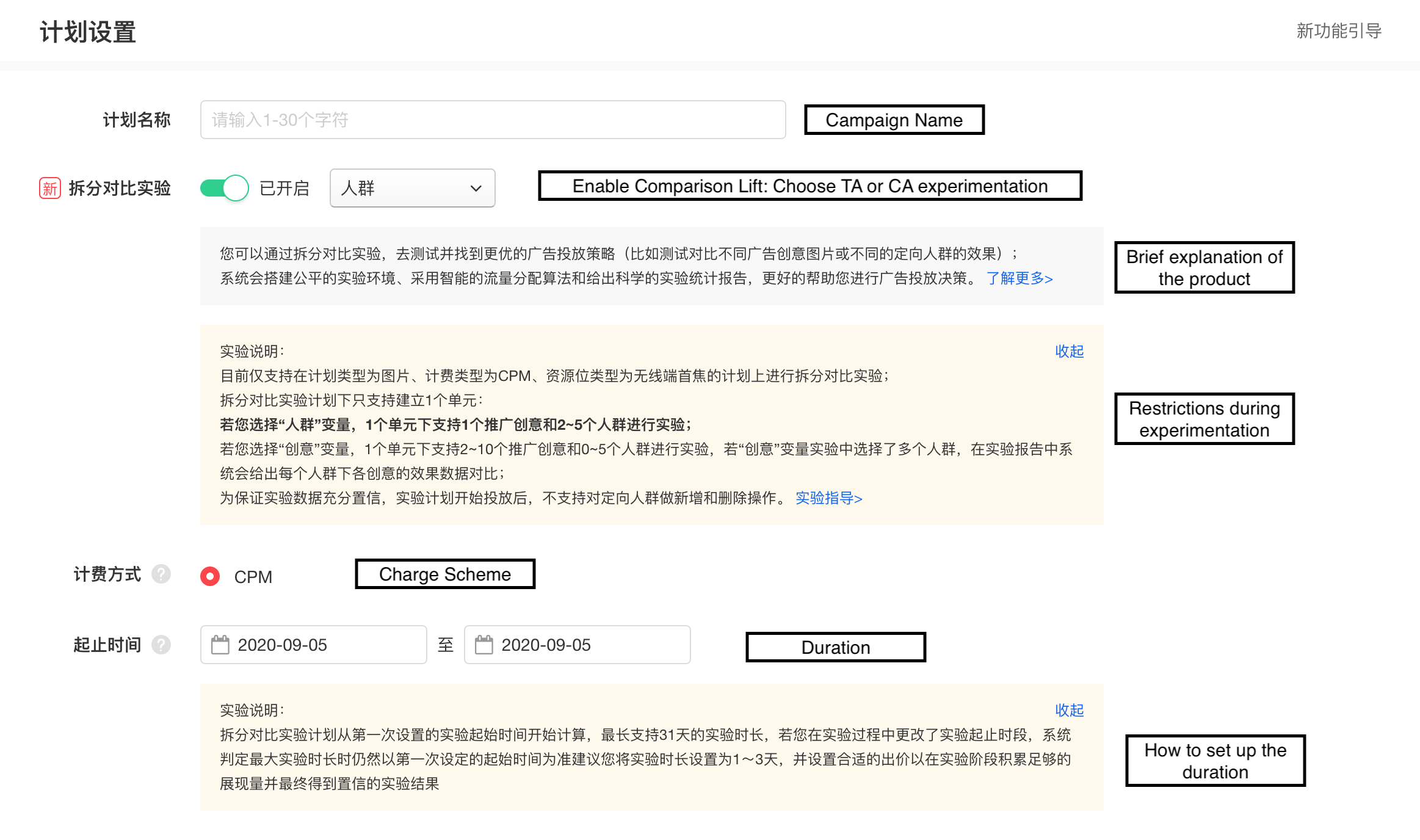}
	\caption{Enabling Product During Campaign Setup}
	\label{fig:exp_setup}
\end{figure}

The UI for \begin{footnotesize}\texttt{Comparison Lift}\end{footnotesize} is built directly into JD's ad-campaign management system. The idea is to make it easy for an advertiser to access the product while she sets up a new campaign and to allow her to easily apply the best discovered options from the experiment to her new campaign. It also has the advantage of making advertisers aware of the availability of EaaS solutions for campaign design. Figure \ref{fig:exp_setup} shows a screenshot of the interface; the green button enables the \begin{footnotesize}\texttt{Comparison Lift}\end{footnotesize} product.

Once the product is enabled, it sets up an experimental ad-campaign on behalf of the advertiser on the JD ad-system. The experimental campaign is similar to a typical ad-campaign, involving rules for bidding, budget, duration etc. The difference is that the advertiser defines $K$ \emph{TA}s and binds $R$ creatives to the experimental-campaign, rather than one as typical; and the allocation of creatives to a user impression is managed by the TS algorithm. Both $K$ and $R$ are limited to a max of 5. Because the algorithm disjoints \emph{TA}s, the number of contexts grows combinatorially as $K$ increases, and this restriction keeps the total combinations manageable. We require the experimental ad-campaign to have the same bids for all compared target audiences and creatives, and all intelligent bidding options turned off; and for the advertiser to refrain from making changes to these settings while the experiment is running.

During the setup, advertisers are asked to whether they wish to run a creative or a target audience experiment (Figure \ref{fig:exp_setup}). This is a matter of terminology. We use the term ``creative experiment'' to refers to experiments that consist of more than 1 creative; otherwise we call them ``target audience experiments.'' Although the underlying algorithm is the same, this distinction helps advertisers navigate the product better, and also affects the specifics of the reports they receive. Once an experimental ad-campaign is setup, its meta-data is passed to a data system described below.

\subsubsection{Data System and Infrastructure}
\begin{figure}
\centering
\includegraphics[width=0.47\textwidth]{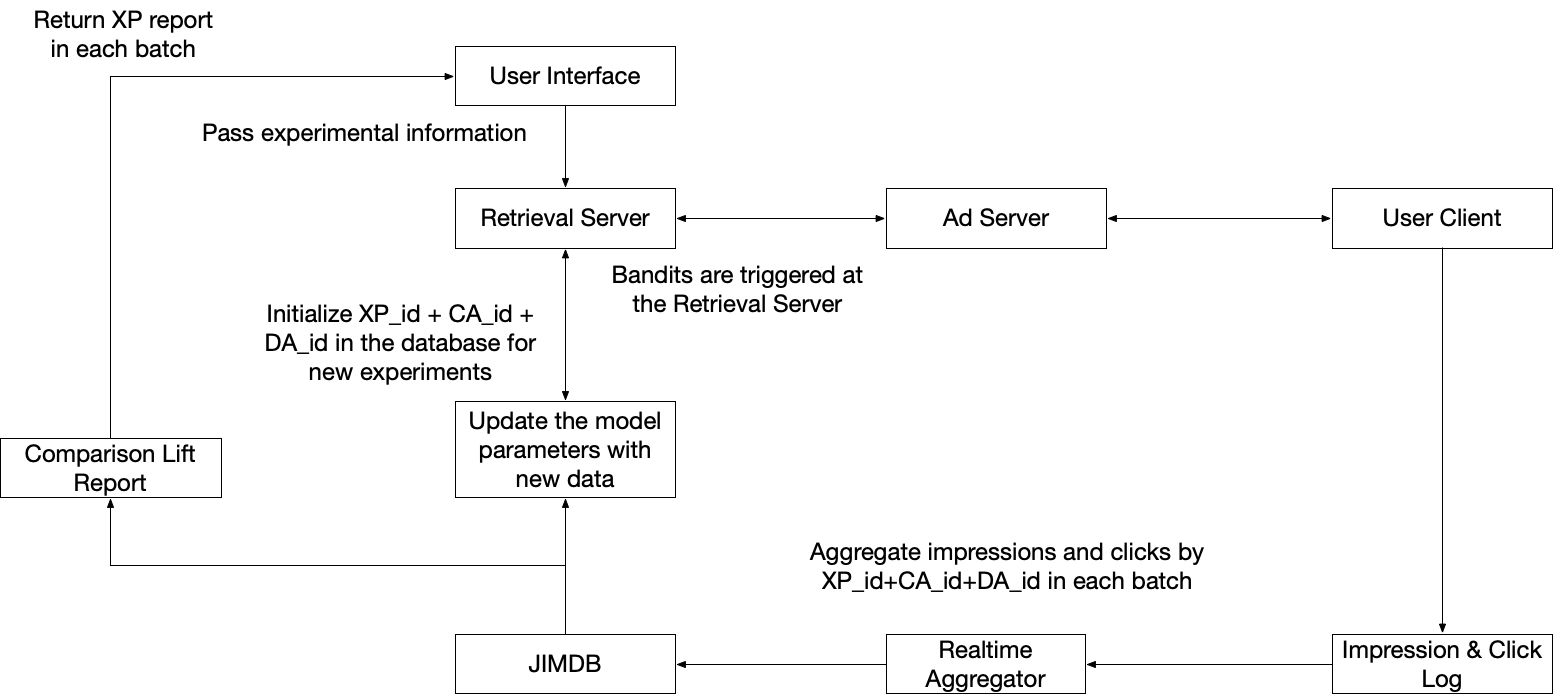}
\caption{Data System and Infrastructure}
\label{fig:infra}
\end{figure}

The first step in the data system is to initializate the experiment in the ad-serving system. To do this, the experimental ad-campaign's meta-data is passed to a Retrieval Server, which maintains a database of all ad-campaigns on the JD ad-system. Here, a campaign record is created for each experimental ad-campaign, and parameters of its bandit are initialized. When a user arrives at an ad-slot, an Ad Server collects the information of the user impression and sends it to the Retrieval Server. The Retrieval Server locates all advertising campaigns that are relevant to the user impression. If an experimental ad-campaign is retrieved, its bandit is triggered, and the optimal creative for it is chosen based on its TS. After this, the Retrieval Server collects the information on the chosen creatives along with the information on all other relevant advertising campaigns and sends it back to the Ad Server. The Ad Server runs an auction, and decides which ad to show to the user. After the ad is served, user outcomes are collected in impression and click logs. After each batch (a batch is typically ten minutes), an Aggregator processes the data in the log-files, and updates the parameters in the Bandit Database. The updated parameters are used to determine the TS in the next batch and to generate a report to the advertisers. Figure \ref{fig:infra} depicts this visually.

\subsubsection{Reporting}
\begin{figure}
	\centering
	\includegraphics[width=0.4\textwidth]{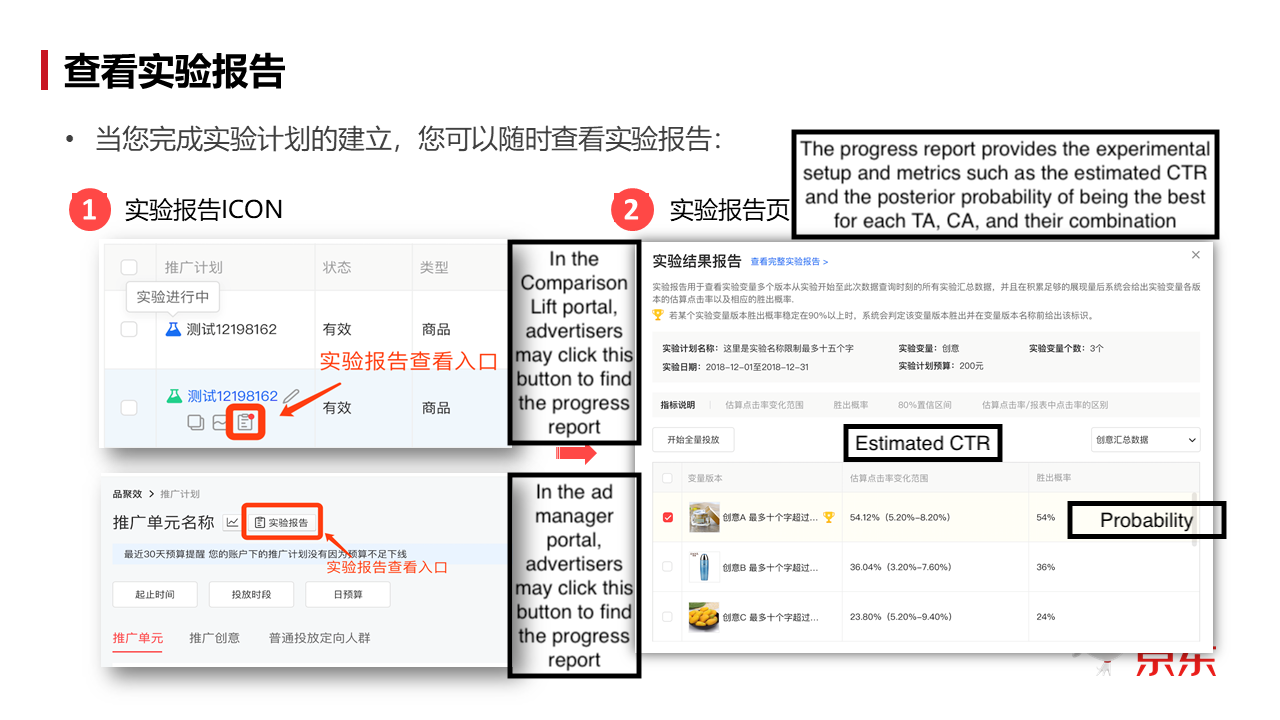}
	\caption{Experimental Report}
	\label{fig:exp_during}
\end{figure}

Reporting is provided via a UI.  The following metrics are displayed: estimated CTRs, credible intervals, and the probability being the best creative, target audience, and its combination (See Figure \ref{fig:exp_during}; numbers in the figure for demonstration only). 

Advertisers can check results from experimental ad-campaigns anytime during the test and performance metrics are calculated based on the TS parameters at the time of checking. In addition, advertisers can also choose to stop and restart the experimental ad-campaign anytime. The Bayesian basis of the algorithm is helpful for providing valid inference in these situations by basing inference on a posterior distribution that conditions on the data collected. Such flexibility is typically not available in other fixed-sample experimental products, and one needs alternative frameworks for frequentist sequential hypothesis testing to accommodate it appropriately (e.g., \cite{johari2015, Juetal2019}. 

When any \emph{C-TA} attains more than 90\% posterior probability of being the best, the advertiser is notified and a final test report generated. This report includes more data about the test such as number of impressions, costs, and other information, besides the estimated CTRs, credible intervals, and the posterior probability of being the best. For target audience experiments, we report the performance of each target audience; while for creative experiments, we report the overall performance of each creative, their performance for each target audience (if more than 1), and the performance of each creative-target audience combination across all combinations.

Even though a final report has been generated at this stage, the advertiser may choose to continue the experimental ad-campaign. Because the product adaptively allocates traffic to best performing creatives, the continuation of the test is aligned with the advertiser's goal of using the best performing variant in her campaigns (a key advantage relative to experimental schemes that impose fixed allocation of traffic across variants). Thus, there is little danger of unnecessarily wasting money or running the test for ``too long.'' An added benefit is the cost to the advertiser of monitoring and leveraging results is also lowered. Finally, after viewing reports, the experimenting advertiser may apply the winning versions to new, regular advertising campaigns with a single click. 

\section{Business Impact and Lessons Learned}

\subsubsection{Product Usage and Value Generated}

\begin{figure}
	\centering
	\includegraphics[width=0.35\textwidth]{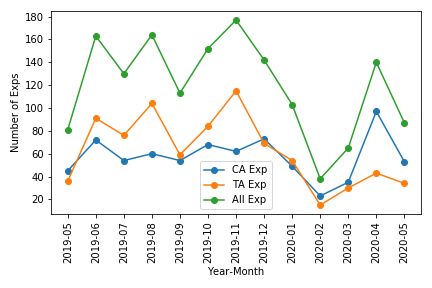}
	\caption{Number of Experiments}
	\label{fig:num_exp}
\end{figure}

One metric of the business impact of the product is its utilization. Between launch in May 2019 and June 2020, advertisers ran 1,547 tests using the product. Figure \ref{fig:num_exp} shows the number of tests by month since launch. The utilization is fairly stable. The drop in February 2020 likely reflects the Chinese Lunar New Year and the initial reaction to the start of the COVID-19 lock-downs in China.

Table \ref{tab:exp_metrics} presents summary statistics for the tests. On average, a test runs for roughly 5.2 days. On average, the number of creatives that are compared is 2.4, and the number of target audiences that are compared is 3.78. Although we set a threshold of 90\% posterior probability to notify and generate a final report, advertisers often end tests before the threshold is met. On average, tests are ended when the maximum posterior probability of being the best option is 73.7\%.

Similar to the previous case study, we compute the value of experimentation and value of adaptive design metrics to illustrate the value generated by the product for advertisers. To assess these metrics credibly, we restrict the computation to tests in which the best option at ending has a posterior probability of being the best that is higher than 90\%. This increases the chance that the best option found in the test will indeed be utilized by the advertiser for subsequent campaigns. Additionally, we drop from the computation, tests in which there are \emph{C-TA} combinations for which there are less than 1,000 impressions. This reduces the statistical error in the CTRs used in simulating the counterfactual scenarios required for the metrics.

\begin{figure}
	\centering
	\includegraphics[width=0.35\textwidth]{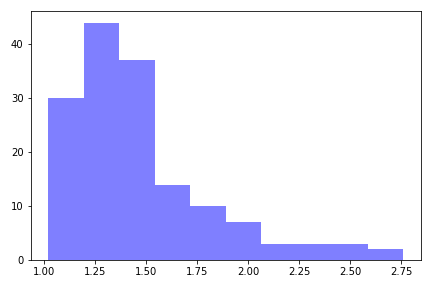}
	\caption{Histogram of Value of Experimentation}
	\label{fig:exp_VOE}
\end{figure}

Figure \ref{fig:exp_VOE} presents a histogram of the value of experimentation metric across this subset. The average is 1.46 (see Table \ref{tab:exp_metrics}). The value can be as high as 2.75 in some tests, indicating substantial uncertainty reduction for advertisers by using the product.

\begin{figure}
	\centering
	\includegraphics[width=0.4\textwidth]{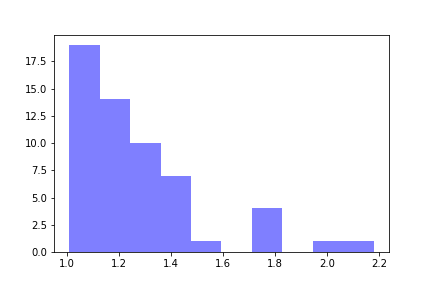}
	\caption{Histogram of Value of Adaptive Design}
	\label{fig:exp_VOB}
\end{figure}

We compute the value of adaptive design only for tests within this subset that have more than 1 creative, since adaptive allocation occurs only in such tests. Figure \ref{fig:exp_VOB} presents a histogram of this metric in this subset. On average, the value is 1.27 and can be greater than 1.6 in certain cases. Clearly, the adaptive design has been able to significantly cut down the opportunity cost of experimentation for the advertisers.

\begin{table}
	\centering \footnotesize
\begin{tabular}{|r|r|}
	\hline 
	Variable (type) & Value \\
	\hline \hline
	No. of Exp. (total) & 1,547 \\
	Creatives in Exp. (mean) & 2.42 \\
	TAs in Exp. (mean) & 3.78  \\
	Exp. No. of Days (mean) & 5.20  \\
	Post-Prob Best Arm at End. (mean) & 73.7\%  \\ \hline
	Value of Exp. Metric (mean) & 1.46 \\
	Value of Adp. Desn. Metric (mean) &  1.27 \\
	\hline
\end{tabular}
\caption{Product Usage Metrics}
\label{tab:exp_metrics}
\end{table}

\subsubsection{Takeaways and Lessons learned}
We detail some summary takeaways from our experience in building and deploying the product. A first point is somewhat obvious in retrospect, but is worth emphasizing because it stands in contrast with some of our own priors going into the product development phase, and also with some lay beliefs we have heard expressed in the academic community. One view of advertiser and marketer behavior is that prior to launching a campaign, advertisers or marketers already have a good understanding of the performance implications of various possible campaign features, and that launched campaigns reflect design features picked optimally based on this knowledge. Under this view, there is limited scope for a product that helps pick good options to meaningfully improve an advertiser's campaign design. A related view is that advertisers also know how much they value the ad-impressions they obtain as part of their marketing campaigns. For instance, a canonical model of an advertiser buying ads on a digital platform is of an agent who knows her valuation, bidding in an auction against her competitors for the impression.   

Our experience has been that neither of these views may have much support in many practical settings. Advertiser uncertainty about the effect of her marketing and advertising campaigns is high, and prior to product development, many advertisers reported launching campaigns with the purpose of discovering best options and resolving this uncertainty. Indeed, one motivation for developing the product was the observation that the same advertiser often launches multiple campaigns on the JD ad-platform with different creatives for the same target audience, or with different target audiences for the same creative, likely reflecting her attempts to discover which works best via quasi-experimentation. Many advertisers also face substantial uncertainty about her own valuation of ad-impressions. The reasons are varied and could be because advertisers' database structure, and statistical, economic sophistication may be poor; because media buying is implemented through a set of intermediaries and not by the advertiser herself; because not enough past historical data has been accumulated for a particular type of audience or media being considered; or because personnel managing at campaigns have left the company without proper knowledge transfer. These aspects imply that an experimentation product of the type described here may be valuable in helping resolve advertiser uncertainty; and that helping the advertiser discover good options in a principled way has the potential to add substantial value by improving campaign performance. 

The second point relates to the appropriate positioning of a measurement product such as \begin{footnotesize}\texttt{Comparison Lift}\end{footnotesize} to obtain business traction. Our experience suggests that its better if measurement is not an end goal in itself, and a measurement product is likely to be utilized more often when it is embedded within the context of a clear decision for the advertiser. For instance, an experimentation system is more likely to be successful as an external-facing business product if it delivers not just causally valid treatment effects for a campaign, but also a clear recommendation for how those treatment effects can be converted into an actionable decision such as bidding, budget allocation, creative optimization or targeting. By that token, we believe the success of \begin{footnotesize}\texttt{Comparison Lift}\end{footnotesize} is linked to the fact that the sophisticated measurement capability it encapsulates leads to a clear and transparent decision (choice of campaign creative and/or target audience), and that this recommendation is made actionable by allowing the advertiser to apply it in a seamless manner to the next campaign she wishes to run. An implication for science is that novel measurement methodology is more likely to have practical impact when developed within the context of a decision-theoretic framework that provides clear policy recommendations. This latter aspect has been emphasized in the literature on ``policy-relevant'' treatment effects and modeling by several thoughtful scholars (e.g., \cite{Heckman2000, Rust2019}), but is worth reiterating.

A third point pertains to the idea of providing experimentation as ``a service'' by the platform to the advertiser via a self-serve automation product that advertisers can access. In this product-flow, the advertiser makes decisions about what she wants to test, while decisions about how the test will be implemented, including sample sizes, traffic allocation to various test and control arms, as well as rules for test stoppage are determined by the platform, acting as an agent of the advertiser. Allowing the platform to run the experiment on behalf of the advertiser in this manner has advantages relative to the advertiser running an experiment on her own. 

One reason, which is elucidated in further detail below, is that the technology platform often has higher statistical sophistication than advertising media buyers, and consequently, the platform is able to better determine aspects that bear on experimental performance such a sample sizes, stopping rules and experimental designs in many situations. A second reason is that modern ad allocation mechanisms, ad-serving and tracking systems are very complex. Digital ad-experiments work well when they incorporate carefully details of how campaign features interact with the ad-allocation mechanism and the ad-serving system in order to obtain exposure, and how cross-device, cross-channel tracking of user behavior is managed and actualized. The platform is in a better position to do this than the advertiser. A final reason is the platform may have better information than the advertiser on details such as a priori expected effect sizes and power because it has access to a much larger corpus of past experiments or historical data it can leverage. This may allow it to do more effective experimental design. For all of these reasons, our view is that platform facilitated experimentation as a service is a better model for ad-experiments on complex publishing platforms rather than advertiser implemented direct experimentation.

\section{Possible extensions}

Several extensions are possible for improving the impact and usability of the product. Possibly the most valuable ones would be to increase its scope in terms of handling comparisons of higher dimensionality and type, and in terms of basing the comparisons on more varied metrics and types of ad-inventory.

On the dimensionality issue, the current version of the product limits the maximum number of treatment arms (creative-audience combinations) to 25. This restriction is driven both by considerations of statistical power in identifying the best arm, as well as concerns about latency in real-time implementation when displaying options from a larger comparison set. As the number of creatives or the number of audiences increase, the number of distinct subpopulations that map to slices of the compared audiences also explode combinatorially, which increases the search complexity for the contextual bandit and increases exploration time. Non-parametric learning approaches may not scale well in such a situation. One possibility is to impose some structure on the nature of payoffs, for instance imposing that the click-through rate has a logistic or probit structure over features, thus projecting the search over parameters on these features and allowing for cross-arm learning, and accelerating identification (e.g., \cite{scott2015multi}). As the exploration phase of the bandit increases, Markov Chain Monte Carlo methods may also become slow because of the need to update the full posterior with data from the beginning of the experiment. Simulated Monte Carlo methods (e.g., \cite{cb2013}) might be attractive in such a situation.

In terms of type of campaign characteristics that are compared, in addition to creatives and target audiences, advertisers seek to understand what bids to choose for various audiences, what channels to target them at, as well as how much budget to allocate across various campaigns. Extending the current product to allow comparisons across these aspects would be valuable. Adaptive bandit-based methods for locating optimal bids in an auction-based advertising environment are discussed for instance in \cite{wncx2019}, and could be adopted for this purpose. Practical optimization methods for automated campaign channel selection and automated budget allocation are discussed for instance in \cite{PaniRaghavanSalin2018}. A product that has a similar optimization flavor such has been developed and deployed at JD.com, though it is not adaptive. Developing an adaptive version of these methods will allow these to be ported to \begin{footnotesize}\texttt{Comparison Lift}\end{footnotesize} experimentation. 

In terms of metrics, the current version of the product optimizes a performance criteria for the advertiser that depends on the CTRs of the options being evaluated. The economic benefit of choosing the option to the advertiser is modeled as a constant times the CTR, where the constant is picked based on historical data outside of the experiment. Extending the performance criteria to other metrics such as conversion or revenue will be valuable, though more challenging. One reason is that conversion is a rarer and more variable event than clicking, therefore more data is required to pick the best performing option under this metric. Another challenge is that clicks encapsulate short-run response to ads and from immediate feedback for the bandit, while conversion and revenue is generated over a longer term and can only provide much more delayed feedback to the bandit. Delayed feedback implies that fast adaptation of experimental design to assessed progress towards the goal is also delayed (see \cite{joulani2013online} for an overview).

One way to address the first issue may be to pool information from past experiments or historical data, or even from models trained on historical data, so as to ``warm-start'' the bandit and to reduce the data requirements from exploration. One way to address the second would be to use surrogates (in addition to clicks) that are observed in high frequency. While the surrogates only proxy for conversion or revenue, they can provide faster feedback and could be the basis of adaptation. Eventually, this approach could also be adapted to accommodate the long-term profit to the advertiser from the evaluated options as a performance metric \cite{athey2016estimating}. The long-term profit is attractive as an economic goal, but is currently difficult to measure reliably unless the experimental intervention is maintained for a long period of time. Efforts along these lines are underway in our team.

In terms of types of ad inventory, the main distinction is between \textit{auction-driven} versus \textit{guaranteed} inventory and between \textit{internal} inventory (such as on the homepage of the JD.com app for which \texttt{JD} is the publisher) and \textit{external} inventory (such as on a \texttt{Toutiao} news feed for which \texttt{JD} serves as the demand side platform or DSP on behalf of the advertiser). The presence of the auction complicates inference with the bandit because the arm that is pulled, a specific creative, may or may not win the auction, and therefore compliance (i.e., ad-exposure) with the treatment arm that is pulled is imperfect, mediated by the auction. This moves statistical inference from a treatment to an intent-to-treat framework, requiring some adaptation of the algorithm and its interpretation relative to that with guaranteed inventory, for which there is no real-time auction. See \cite{linnairsahniwaisman2019parallel} for some ways this is handled in a non-adaptive setting; extending this to an adaptive setting is a work in progress. The implication of external versus internal inventory is that as the publisher, JD.com observes the auction-queue and the winner of the auction on internal inventory, but does not observe the auction queue, or the winner when it loses the auction on external inventory. This means that for internal inventory, we can leverage knowledge of the auction-queue and the winner for improving the regret performance of the algorithm as well as improving statistical power, neither of which is available for external inventory. Developing ways of extending \begin{footnotesize}\texttt{Comparison Lift}\end{footnotesize} to these situations is again a work in progress.

Other extensions pertain to the human element associated with the product's business use. While there are pockets of sophistication, by and large, advertiser knowledge of sequential experimentation, statistical methodologies and causal inference and interpretation is limited compared to the data science community. Several advertisers frequently ``peek'' at the results from the tests while the experiment is running, and terminate the experiment before the criteria specified in the stopping rule is met. Some advertisers run tests with insufficient budgets, causing their experiments to stop too early. Both result in suboptimal performance because the bandit does insufficient exploration and has not had the opportunity to discover the true-best arm at the forced stopping point. More statistical knowledge amongst users of the product and amongst marketing media buyers can help address the early stopping issue perhaps facilitated by training from the platform itself. This could increase advertisers awareness of formal statistical procedures and increase advertiser willingness-to-pay for experimentation. Better assessment of statistical sample size requirements and experimentation costs by the platform can provide advertisers more accurate suggestions of appropriate levels of budgets to allocate for their tests, which could address the budget exhaustion issue.

Finally, a related extension is linked to the observation that in comparing options using the product, several advertisers vary many factors and levels at the same time. For instance the creatives that are compared may include one featuring a picture of the product along with a particular promoted price, and another with no prices, but picturing a customer using the product. \begin{footnotesize}\texttt{Comparison Lift}\end{footnotesize} is able to tell the advertiser which one works better, but running one-off comparisons in this manner may not be helpful in programmatically building a knowledge-base for the advertiser. For instance, if we find in the above example that the first creative is better, it is not clear whether it is due to the fact that displaying prices is good relative to not displaying, or whether showing product images is better than showing customer images. To build knowledge programmatically, it might be better for the advertiser to vary different levels of one factor across different comparisons while holding other factors fixed, and to do this in a systematic way across considered factors. 

\begin{figure}
	\centering
	\includegraphics[width=0.3\textwidth]{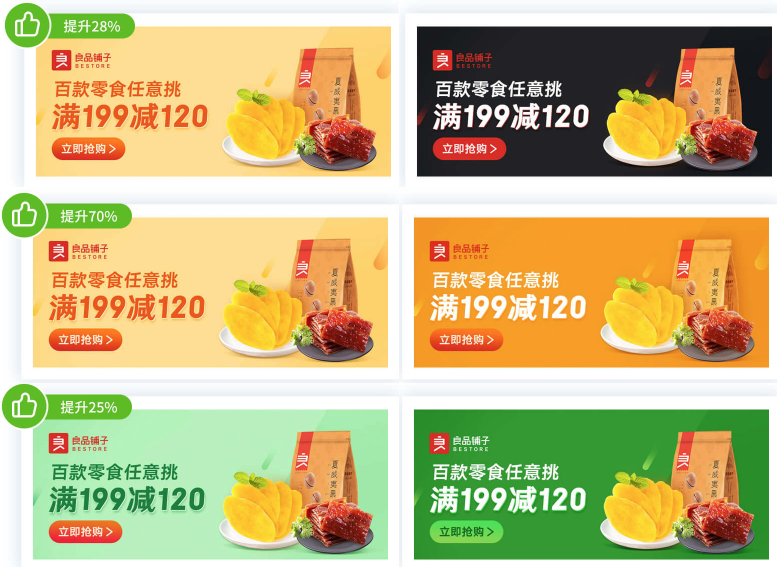}
	\caption{Example of a Testing Plan}
	\label{fig:test_plan}
\end{figure}

One way to achieve this would be to build awareness about experimental design in the advertiser community by facilitating training. Another way would be to extend the product so as to elicit from the advertiser a set of learning goals, and to suggest to her a particular sequence of options to compare that is based on those goals. This method has been implemented at the company by allowing the experimenting advertiser to collaborate with JD's internal design studio, which designs a suggested series of creatives for the advertiser that can then be tested via \begin{footnotesize}\texttt{Comparison Lift}\end{footnotesize}. For example, in Figure \ref{fig:test_plan}, the studio helped the advertiser design creatives with the same content but different background colors, allowing the testing to identify which colors produce superior CTRs, holding other aspects fixed. More case-studies are available at \url{https://ling.jd.com/cms/page/ctr618?ADTAG=618.ctr.erpbanner}.

\section{Conclusions}
Experimentation is a powerful tool to credibly evaluate possibilities and to discover good options in the presence of uncertainty. In offline settings, experimentation has traditionally been hard to implement and costly to induce. A distinguishing feature of modern online advertising markets is the ease of experimentation facilitated in many cases by the publishing platform. EaaS products such as the ones discussed here make experimentation available ``on-tap'' to digital marketers for improving their campaign design and optimization. Given the significant value they can generate, such products are expected to be a key part of advertising platforms' product portfolios and firms' marketing campaign planning toolkits, going forward. 

\section*{Acknowledgments}
We thank Jack Lin, Paul Yan, Lei Wu and the JD.com engineering, product and sales teams for collaboration and support.

\bibliography{MAB.bib}

\begin{thebibliography}{13}
\providecommand{\natexlab}[1]{#1}
\providecommand{\url}[1]{\texttt{#1}}
\providecommand{\urlprefix}{URL }
\expandafter\ifx\csname urlstyle\endcsname\relax
  \providecommand{\doi}[1]{doi:\discretionary{}{}{}#1}\else
  \providecommand{\doi}{doi:\discretionary{}{}{}\begingroup
  \urlstyle{rm}\Url}\fi

\bibitem[{Athey et~al.(2016)Athey, Chetty, Imbens, and
  Kang}]{athey2016estimating}
Athey, S.; Chetty, R.; Imbens, G.; and Kang, H. 2016.
\newblock Estimating Treatment Effects using Multiple Surrogates: The Role of
  the Surrogate Score and the Surrogate Index.
\newblock \emph{arXiv:1603.09326} .

\bibitem[{Cherkassky and Bornn(2013)}]{cb2013}
Cherkassky, M.; and Bornn, L. 2013.
\newblock Sequential {M}onte {C}arlo bandits.
\newblock \emph{arXiv:1310.1404} .

\bibitem[{Geng, Lin, and Nair(2020)}]{Geng2020}
Geng, T.; Lin, X.; and Nair, H.~S. 2020.
\newblock Online Evaluation of Audiences for Targeted Advertising via Bandit
  Experiments.
\newblock In \emph{AAAI 2020}, 13273--13279.

\bibitem[{Heckman(2000)}]{Heckman2000}
Heckman, J.~J. 2000.
\newblock Causal Parameters and Policy Analysis in Economics: A Twentieth
  Century Retrospective.
\newblock \emph{Quarterly Journal of Economics} 115(1): 45--97.

\bibitem[{Johari, Pekelis, and Walsh(2016)}]{johari2015}
Johari, R.; Pekelis, L.; and Walsh, D.~J. 2016.
\newblock Always valid inference: Bringing sequential analysis to {A/B}
  testing.
\newblock \emph{arXiv:1512.04922} .

\bibitem[{Joulani, György, and Szepesvári(2013)}]{joulani2013online}
Joulani, P.; György, A.; and Szepesvári, C. 2013.
\newblock Online Learning under Delayed Feedback.
\newblock \emph{arXiv:1306.0686} .

\bibitem[{Ju et~al.(2019)Ju, Hu, Henderson, and Hong}]{Juetal2019}
Ju, N.; Hu, D.; Henderson, A.; and Hong, L. 2019.
\newblock A sequential test for selecting the better variant: Online {A/B}
  testing, adaptive allocation, and continuous monitoring.
\newblock In \emph{WSDM 2019}, 492--500.

\bibitem[{Lin et~al.(2019)Lin, Nair, Sahni, and
  Waisman}]{linnairsahniwaisman2019parallel}
Lin, X.; Nair, H.; Sahni, N.; and Waisman, C. 2019.
\newblock Parallel Experimentation in a Competitive Advertising Marketplace.
\newblock \emph{arXiv:1903.11198} .

\bibitem[{Pani, Raghavan, and Sahin(2018)}]{PaniRaghavanSalin2018}
Pani, A.; Raghavan, S.; and Sahin, M. 2018.
\newblock Large-Scale Advertising Portfolio Optimization in Online Marketing.
\newblock \emph{Working Paper, Univ. of Maryland} .

\bibitem[{Russo et~al.(2018)Russo, Van~Roy, Kazerouni, Osband, and
  Wen}]{russo2018}
Russo, D.~J.; Van~Roy, B.; Kazerouni, A.; Osband, I.; and Wen, Z. 2018.
\newblock A tutorial on {Thompson} sampling.
\newblock \emph{Foundations and Trends{\textregistered} in Machine Learning}
  11(1): 1--96.

\bibitem[{Rust(2019)}]{Rust2019}
Rust, J. 2019.
\newblock Has Dynamic Programming Improved Decision Making?
\newblock \emph{Annual Review of Economics} 11(1): 833--858.

\bibitem[{Scott(2015)}]{scott2015multi}
Scott, S.~L. 2015.
\newblock Multi-armed bandit experiments in the online service economy.
\newblock \emph{Applied Stochastic Models in Business and Industry} 31(1):
  37--45.

\bibitem[{Waisman et~al.(2019)Waisman, Nair, Carrion, and Xu}]{wncx2019}
Waisman, C.; Nair, H.~S.; Carrion, C.; and Xu, N. 2019.
\newblock Online inference for advertising auctions.
\newblock \emph{arXiv:1908.08600} .

\end{thebibliography}
\end{document}